\documentclass[manuscript]{acmart}
\AtBeginDocument{%
  }

\setcopyright{acmlicensed}
\copyrightyear{2018}
\acmYear{2018}
\acmDOI{XXXXXXX.XXXXXXX}
\acmISBN{978-1-4503-XXXX-X/2018/06}

\usepackage{subcaption}
\usepackage{multirow}

\begin{document}

\title{Protecting and Preserving Protest Dynamics for Responsible Analysis}


\author{Cohen Archbold}
\email{cohen.archbold@uky.edu}
\orcid{0009-0003-1275-8329}
\author{Usman Hassan}
\email{usman.hassan@uky.edu}
\orcid{}
\author{Nazmus Sakib}
\email{}
\orcid{}
\author{Sen-ching Cheung}
\email{sen-ching.cheung@uky.edu}
\orcid{}
\author{Abdullah-Al-Zubaer Imran}
\email{aimran@uky.edu}
\orcid{}

\affiliation{%
  \institution{University of Kentucky}
  \city{Lexington}
  \state{Kentucky}
  \country{USA}
}


\begin{abstract}
Protest-related social media data are both valuable for understanding collective action and inherently high-risk due to the ethical concerns surrounding surveillance, repression, and individual privacy. Contemporary AI systems can identify individuals, infer sensitive attributes, and cross-reference visual information across platforms, enabling forms of surveillance that pose significant risks to protesters and bystanders. In such contexts, large foundation models trained on protest imagery risk memorizing and disclosing sensitive information about individuals identities, locations, and actions, which can lead to to cross-platform identity leakage, retroactive participant identification, and state or institutional surveillance. Despite growing interest in automated protest analysis, existing approaches do not provide a holistic pipeline that jointly integrates privacy risk assessment, downstream analysis, and fairness considerations. To address this gap, we propose a responsible computing framework designed to support the analysis of collective protest dynamics while intentionally reducing risks to individual privacy. Our framework replaces sensitive protest imagery with well-labeled synthetic reproductions using conditional image synthesis, enabling analysis of collective patterns without direct exposure of identifiable individuals. Through extensive evaluation, we demonstrate that our proposed approach produces realistic and diverse synthetic imagery while balancing downstream analytical utility with meaningful reductions in privacy risk. We further assess demographic fairness in the generated data, examining whether synthetic representations disproportionately affect specific subgroups and thereby situating equity as a core component of responsible protest analytics. Rather than offering absolute privacy guarantees, our method adopts a pragmatic, harm-mitigating approach that illustrates how responsible design choices can enable socially sensitive analysis while acknowledging residual risks and misuse potential.

\end{abstract}




\maketitle

\section{Introduction}
Protests are a complex and dynamic expression of social sentiment, often sparked by perceived injustices. According to the 2020 World Peace Index by the Institute for Economics and Peace, there was a 102\% increase in protests, riots, and strikes worldwide between 2011 and 2018. Among these protests, nonviolent demonstrations accounted for 64\%, general strikes for 6\%, and riots for the remaining 30\%~\cite{iep2020}. Social media data presents a new avenue to view and analyze these events in real-time, providing information to protesters, lawmakers, and policymakers. In the visual data domain, social event imagery has previously been analyzed to detect activities and estimate violence~\cite{UCLA_protest}, to estimate protest size~\cite{steinert2022violence}, to assess the correlation between the severity of state repression~\cite{steinert2022violence} and changes in subsequent protest size, to estimate protest topics~\cite{casm2019}, to estimate demographic information (e.g., age, gender, and race)~\cite{UCLA_protest}. These works demonstrate the technical feasibility of automated protest analysis and its potential to inform public discourse and policy.

However, these analyses raise significant ethical and legal concerns when applied to politically sensitive populations. Individuals captured in protest imagery are often engaged in acts of political expression, making them particularly vulnerable to harmful downstream use or misinterpretation in the absence of appropriate safeguards. Furthermore, protest images frequently capture large crowds and complex interactions, often including individuals of diverse races, ages, and genders—thereby increasing the likelihood of incidental exposure of bystanders or unintended contextual interpretations. Recent data protection regulations, such as the General Data Protection Regulation (GDPR) \cite{gdpr2016} and the California Consumer Privacy Act (CCPA) \cite{ccpa2018}, impose strict requirements on the collection, storage, and use of personal data. However, legal definitions of personally identifiable information do not necessarily align with practical risks of re-identification or downstream misuse, especially in visual data. As a result, compliance with data protection regulations alone does not guarantee responsible use in socially sensitive contexts, motivating technical approaches that explicitly account for ethical risk and potential misuse. 


One prominent class of such approaches seeks to reduce exposure to sensitive personal information at the data level. Synthetic data has emerged as a promising approach to mitigate aforementioned risks by generating imagery that preserves utility while removing identifiable information \cite{lin2022privacy, faisal2022medical}.  \citet{sun2023gan} generate synthetic magnetic resonance images (MRIs) of vertebral units (VUs) to support medical data sharing. Similarly, \citet{xioing2019priauto} use a Generative Adversarial Network (GAN) to synthesize and protect vehicular driving data. Yet, whether synthetic data fully complies with personal data regulations remains unsettled~\cite{boudewijn2024legal}, particularly in high-stakes domains where even partial re-identification can lead to serious harm. Despite the significance of these risks, existing image-based protest analysis techniques do not jointly address (1) the privacy risks associated with analyzing and releasing protest imagery and (2) the potential biases that may arise in model predictions or datasets. 
Our proposed method aims to fill this gap by enabling protest imagery analysis while reducing risk to individual privacy and enabling group fairness accounting in the evaluation process.


Our solution provides a framework for conducting image-based analysis in socially sensitive datasets while reducing exposure of identifiable individuals. Formal privacy mechanisms such as Differential Privacy offer strong theoretical guarantees but remain challenging to apply to large-scale visual models due to scalability constraints, complex data distributions, and substantial privacy–utility trade-offs~\cite{yoon2018pategan, long2021gpate, harder2023pretrained}. We therefore explore synthetic data generation as a potential approach for mitigating privacy risk in deep learning pipelines by replacing sensitive raw data with generated imagery for task-oriented training; however, the extent to which such synthetic data meaningfully reduces privacy risk remains an open question. While generative models are not designed to provide formal privacy guarantees, prior work suggests that they can offer a bounded level of protection when used as a data source~\cite{Chueng2018sensitive,lin2022privacy}. Beyond protecting privacy, our framework also incorporates a structured evaluation of fairness and potential biases in the synthetic imagery—allowing us to assess whether sensitive attributes such as race, age, or gender are disproportionately represented in ways that could influence downstream model performance. Our goal is not to enable unrestricted deployment, but to provide an evaluable framework that supports aggregate-level protest analysis while making privacy–fairness trade-offs explicit.

In this study we propose a generative method for privacy-aware protest analysis. We explore both formal privacy mechanisms and non-private generative models, assessing the utility, fairness, and privacy implications of using synthetic data for downstream tasks. Our approach frames protest analysis as a responsibility-sensitive task and evaluates design choices through the lenses of privacy, utility, and fairness.
Our key contributions include:
\begin{enumerate}
    \item A responsibility-driven framework for image-based protest analysis that prioritizes privacy protection and harm mitigation in socially sensitive data;
    \item A conditional image synthesis approach that generates high-fidelity and diverse protest imagery while reducing reliance on identifiable original data during model training;
    \item A comprehensive evaluation of privacy–utility trade-offs for multi-level protest image understanding tasks;
    \item A systematic fairness analysis assessing demographic representation and its impact on downstream model behavior.
\end{enumerate}

\section{Related Work}
\subsection{Protest Analysis}

Social dynamics across domains such as social networks, geographic data analysis, and information retrieval have been extensively studied~\cite{Korkmaz2016MultisourceMF,Kharroub2016SocialMA, embers, Alvi2023OnTF}. Significant efforts have focused on collecting and categorizing multilingual social media data, leading to numerous studies predicting civil unrest~\cite{Korkmaz2016MultisourceMF,casm2019}, analyzing information warfare, and forecasting election outcomes~\cite{Ramteke2016ElectionRP, Alvi2023OnTF}. While existing methodologies have advanced modeling social movements from textual and geographic perspectives, comparatively less attention has been given to visual analysis of protests, especially in their potential impact on privacy and bias. Many early works in the image domain focus on specific movements, often constrained to one hashtag or geo-location. \citet{Kharroub2016SocialMA}  catalog and contrasts the visual features of imagery from the 2011 Egyptian revolution. \citet{blm_computervision} analyze Facebook imagery to study visual characteristics of the Black Lives Matter movement. Recent advancements in data collection have enabled the application of deep neural networks to classify protest imagery and analyze its visual features~\cite{casm2019, UCLA_protest}. The work by \citet{UCLA_protest} on analyzing protest violence and sentiment provides a valuable foundation for understanding the dynamics of protests through image analysis. In their approach, the authors collect a large protest image dataset and annotate it for protest presence, violence, visual attributes, and sentiment using crowdsourced human annotators. They further introduce a benchmark dataset and a deep neural network for modeling these annotations, which we adopt as a baseline for our analysis. Lastly, these works primarily focus on predictive performance and dataset construction, with limited consideration of privacy risk or fairness in visual protest analysis.

\subsection{Privacy Risks in Machine Learning}
The widespread deployment of computer vision models for image analysis has raised substantial concerns regarding privacy risks in visual data. Public-facing models can inadvertently memorize and leak sensitive information about individuals captured in their training sets~\cite{shokri2017membership,salem2018mlleaks,DBLP:journals/popets/HayesMDC19,rezaei2021difficulty}. In the sensitive domain of protest analysis, these vulnerabilities are not merely theoretical; they represent a bridge between digital data and physical risk. By enabling the deanonymization of protesters or the extraction of protected identities, these attacks can be weaponized for targeted surveillance and the suppression of civil liberties. These vulnerabilities have motivated a range of adversarial techniques in which attackers exploit model outputs or internal states to extract private information. These include:

\begin{enumerate}
    \item \emph{Membership Inference Attacks}, which attempt to determine whether a specific individual’s data was included in a model’s training set~\cite{shokri2017membership, salem2018mlleaks, DBLP:journals/popets/HayesMDC19, rezaei2021difficulty}.
    \item \emph{Attribute Inference Attacks}, which seek to uncover sensitive attributes like race, age, or gender from model outputs or embeddings, even when such attributes were not explicitly used as prediction targets~\cite{jia2020attriguard, jayaraman2022attribute, struppek2023class}.
    \item \emph{Reconstruction Attacks}, which can partially or fully reconstruct original training samples from model gradients or outputs, threatening direct privacy leakage~\cite{zhang2020secret, nguyen2023rethinking}.
\end{enumerate}
Notably, several of these attacks remain effective even when models are trained with formal privacy-preserving mechanisms such as differential privacy~\cite{zhang2020secret,rezaei2021difficulty,carlini2023extracting}, highlighting the limitations of relying solely on theoretical guarantees in complex visual domains. Consequently, these techniques—particularly membership inference attacks—are widely used as practical tools for empirically assessing privacy leakage and auditing model behavior~\cite{jagielski2020auditing,lu2022general,Izzo2022ProvableMI,andrew2024oneshot}. As deep learning systems are increasingly applied to sensitive visual data, accounting for such privacy risks is essential for responsible model development and deployment.

\subsection{Private Image Synthesis}
Generative models like Generative Adversarial Networks (GANs), which use adversarial training between a generator and discriminator, are well-established methods for producing realistic images from noise or semantic inputs. GANs have a rich history in synthetic image generation~\cite{CycleGAN2017,pix2pix2017,karras2019stylebased,karras2021aliasfree}, serving as a benchmark for producing high-quality, realistic visuals across various domains.  StyleGAN~\cite{karras2019stylebased, karras2020analyzing,karras2021aliasfree}, an extension of GANs, enhances control over the visual attributes of generated images by disentangling factors of variation such as pose, identity, and background, capabilities that are particularly relevant in complex crowd and protest scenes. Synthetic images generated by GANs have been investigated for their privacy risk mitigating potential in domains such as autonomous driving~\cite{xioing2019priauto} and medical imaging~\cite{faisal2022medical, sun2023gan}, where they are used as replacements for or supplements to real data. Additionally, GANs have been adapted to employ differential privacy (DP), aiming to provide formal privacy guarantees using obfuscation methods such as noise addition and output aggregation~\cite{xie2018differentially,yoon2018pategan,usman2023hegan,long2021gpate,harder2023pretrained}. However, the effectiveness of both standard and differentially private GANs for complex protest imagery remains unclear, especially in terms of whether generated images may reproduce identifiable individuals, protest locations, or contextual cues that enable re-identification or downstream misuse.

\subsection{Differential Privacy (DP)}
\label{sec:diffpriv}
Let $\mathcal{M}$ be a randomized algorithm with output set $\mathcal{S}$ and neighboring input datasets $X_1$ and $X_2$ that differ by one sample, the formal definition of $(\epsilon, \delta)$-differential privacy is:
\begin{equation}
    \Pr[\mathcal{M}(X_1) \in \mathcal{S}] \leq e^\varepsilon \cdot \Pr[\mathcal{M}(X_2) \in \mathcal{S}] + \delta,
    \label{eq:diff_p}
\end{equation}
where $\varepsilon$ is the privacy level (maximum difference in output distributions), and $\delta$ is the privacy leakage probability \cite{Dwork2014TheAF}. This formulation guarantees that the output of $\mathcal{M}(X_1)$ is (almost) equally likely to occur for any neighboring dataset $\mathcal{M}(X_2)$, limiting the information gained about any single data point. DP-GAN~\cite{xie2018differentially} extends the standard GAN framework~\citep{goodfellow2014generative} by incorporating differentially private stochastic gradient descent. Subsequent DP generative models employ techniques such as teacher ensembles~\cite{long2021gpate}, specialized sampling strategies~\cite{usman2023hegan}, and mean kernel embeddings~\cite{harder2021dpmerf} to improve generation quality. Despite these advances, DP generative models still struggle to capture complex, high-resolution imagery due to substantial learning constraints~\cite{xie2018differentially,yoon2018pategan,usman2023hegan,long2021gpate,harder2023pretrained}. In socially sensitive settings such as protest imagery, these limitations are especially consequential: overly degraded representations can undermine the validity of downstream analysis, while residual leakage risks may still enable harmful inference. These challenges motivate the exploration of alternative, risk-mitigating strategies such as synthetic data generation, while explicitly acknowledging the trade-offs between formal guarantees, practical utility, and social harm.

\subsection{Anonymization}
Instead of leveraging differential privacy, many works attempt to anonymize datasets using various methods to obfuscate sensitive features within imagery, including face and text blurring \cite{mirjalili2018semiadversarial, Mirjalili_2020}, face synthesis \cite{hukkelås2019deepprivacy}, and full-body synthesis \cite{hukkelas23DP2}. While these approaches can offer favorable privacy–utility trade-offs for specific tasks by retaining real image pixels, many identifying features often remain unaltered, including background location, human activities, and textual content. As a result, approaches that operate on limited regions of an image may provide only partial protection. Whole-image synthesis should therefore be explored as a more holistic strategy for reducing exposure of potentially sensitive features.

\subsection{Bias in ML}
Machine learning systems often inherit and amplify biases from their training data~\cite{Mehrabi2019ASO}, a concern that is particularly pronounced in generative models replicating complex data distributions. GANs, in particular, may exacerbate bias due to mode collapse and objectives that prioritize visual realism over equitable representation~\cite{Mehrabi2019ASO, Kenfack2021OnTF}. \citet{Kenfack2021OnTF} find that conditional GANs trained on imbalanced facial datasets tend to under-represent minority identities, amplifying existing power imbalances and risks of harm in socially sensitive domains like protest imagery. To address these issues, recent work has explored strategies including fairness-aware training, latent space interventions, and prompt-based conditioning~\cite{Xu2018FairGANFG,shen2024finetuning}. However, the interaction between fairness and privacy remains unclear, with ongoing debate over whether these objectives are aligned or inherently in tension~\cite{dwork2012fairness,kamath2025bias}. Balancing these objectives remains a critical challenge, particularly as generative models are deployed in ethically sensitive contexts.

\section{Methodology}
\label{sec:methods}

\subsection{Image Generation}
\label{subsec:img_gen}
We generate samples within a conditional framework, utilizing the annotations for each image. This framework models the joint distribution of images and their annotations, enabling controlled generation of synthetic images while preserving attribute-level information. We evaluate the learned conditional distribution using downstream classification tasks, assessing both utility and alignment with privacy-aware objectives.

\begin{figure*}[t]
    \centering
    \includegraphics[width=0.9\linewidth, trim={0.5cm 0.4cm 0.75cm 0.48cm}, clip]{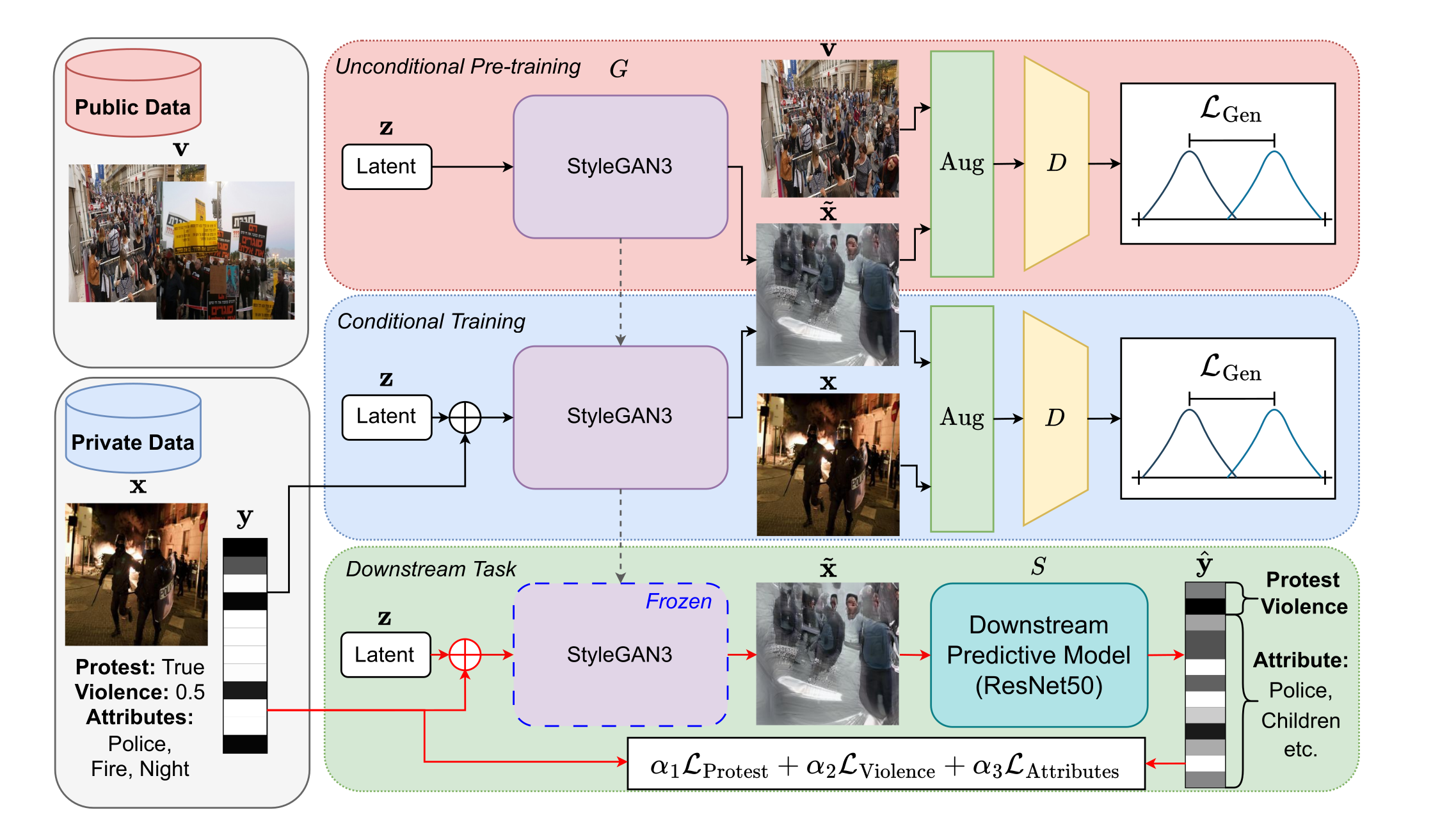}
    \caption{Diagram of our proposed protest analysis framework. We synthesize image $\tilde{\mathbf{x}}$, from the real data annotations $\mathbf{y}$, for training our downstream analysis model privately. The the information flow for the downstream analysis is shown in red.}
    \label{fig:main}
    \Description{Diagram of the information flow of the model.}
\end{figure*}

Our proposed framework adapts StyleGAN3~\cite{karras2021aliasfree} to the challenges of protest imagery through a tailored conditioning strategy and a two-stage training procedure designed specifically for low-resource and highly imbalanced data regimes. Unlike standard StyleGAN training pipelines, which assume large, balanced datasets and focus primarily on visual fidelity, our approach first performs unconditional pretraining on a diverse collection of human-centric imagery drawn from crowd counting and protest-related domains. This stage enables the generator to learn generalizable human and crowd structure without relying on sensitive or sparsely labeled protest data. We then fine-tune the model on the well-annotated dataset of \citet{UCLA_protest}, using conditioning to preserve task-relevant attributes under severe class imbalance. This design allows the model to generate high-fidelity synthetic protest imagery while maintaining control over semantic attributes required for downstream privacy-aware and fairness-focused evaluation. The complete model architecture is illustrated in Fig \ref{fig:main}.

We denote generator network $G$, discriminator $D$, and their respective model weights as $\theta_G$ and $\theta_D$. Given random latent vector $\mathbf{z}$, real image $\mathbf{x}$ and its corresponding annotation $\mathbf{y}$, we train $G$ and $D$ in an adversarial game generating a mini-batch of data points $G(\mathbf{z},\mathbf{y})=\tilde{\mathbf{x}}$ and discriminating real from synthetic samples, $D(\mathbf{x})$ and $D(\tilde{\mathbf{x}})$. We optimize StyleGAN3 \cite{karras2021aliasfree} parameters using the WGAN Loss \cite{arjovsky2017wasserstein} and R1 Regularization \cite{Mescheder2018WhichTM} parameterized by $\gamma$:
\begin{equation}
   \mathcal{L}_{\text{Gen}}(\theta_G, \theta_D) = \mathbb{E}_{\tilde{\mathbf{x}} \sim \mathbb{P}_{g}}\left[D\left(G(\mathbf{\mathbf{z}, \mathbf{y})}\right)\right] - \mathbb{E}_{\mathbf{x} \sim \mathbb{P}_{r}}\left[D\left(\mathbf{x}\right)\right] + \\ \frac{\gamma}{2} \mathbb{E}_{\mathbf{x} \sim \mathbb{P}_{r}}\left[||\nabla_\mathbf{x}{D\left(\mathbf{x}\right)}||^{2}\right], \label{eq:gen}
\end{equation}
where $\mathbb{P}_r$ and $\mathbb{P}_g$ are the distributions of real and synthetic data samples respectively.
This regularization scheme focuses on the discriminator's ability to classify the real data $\mathbf{x}$, reduces sharp gradient changes and enforcing a Lipschitz constraint. After learning the conditional image generation, we freeze the generator and generate synthetic data pairs $(\tilde{\mathbf{x}},\mathbf{y})$, which can be used for downstream tasks while reducing reliance on sensitive real images.

\subsection{Privacy Considerations of StyleGAN}
Due to the degradation in image quality and the challenges of training GANs with differential privacy mechanisms, we adopt StyleGAN without any privacy-specific modifications, and experimentally demonstrate that, with careful training, it can achieve a practical balance between maintaining image fidelity and providing privacy risk mitigation. \citet{lin2022privacy} show that if a GAN is trained (without any special DP mechanisms) on $m$ samples and used to generate $n$ samples, the generated samples satisfy $(\varepsilon, \delta)$-differentially privacy where $\delta = \frac{\mathcal{O}(n/m)}{\varepsilon(1-e^{-\varepsilon})}$ given any $\varepsilon>0$. These results indicate that GAN-generated samples exhibit a form of inherent, albeit weak, differential privacy. In practice, formal DP mechanisms significantly hinder training stability and image quality for high-resolution, complex scenes, making them impractical for protest imagery. We therefore treat StyleGAN as a heuristic tool for mitigating privacy exposure, not as a formally private mechanism. Our approach does not provide differential privacy guarantees, and privacy risk increases with the number of generated samples. This limitation underscores the need for empirical privacy auditing and cautious deployment in socially sensitive settings.


\subsection{Downstream Multilevel Protest Analysis}
\label{subsec:downstream}
For the downstream tasks, including protest detection, protest attribute classification, and violence assessment, we adapt the method of \citet{UCLA_protest} using a deep convolutional neural network to learn the label distribution of synthetic imagery $\tilde{\mathbf{x}}$. These tasks are chosen because they reflect common forms of protest analysis used in practice and allow evaluation of privacy–utility trade-offs at multiple semantic levels. We use the benchmark dataset from \citet{UCLA_protest}, utilizing only the image annotations as input. Given binary labels for protest $y_p \in \{0, 1\}$ and attributes $y_{a,j} \in \{0,yl (1\} $ where $ j\in \{1,2,3,...10\}$ corresponds to the ten annotated attributes provided in the original dataset, we train the network to correctly match the synthetic image $\tilde{\mathbf{x}}$ with the input protest class $\hat{y}_p$ and attribute classes $\hat{y}_{a,j}$. For violence, we predict the continuous labels $v \in \mathbb{R}, v \in [0, 1]$ as $\hat{v}$ that estimates the overall violence level. Consequently, the downstream model $S$, with weights $\theta_S$, produces a vector of predictions $S(\tilde{\mathbf{x}}) = \hat{\mathbf{y}} = [\hat{y}_p, \hat{v}, \hat{y}_{a,j}]$. We follow the proposed loss in \cite{UCLA_protest} for optimization, using a hybrid of cross-entropy loss for protest and attributes, and mean-squared-error loss for violence. Given a mini-batch of $N$ true labels $y$ and predicted score $\hat{y}$ the cross entropy loss for protest is given by,
\begin{equation}
    \mathcal{L}_{\text{Protest}}(\theta_S) = -\frac{1}{N} \sum_{i=1}^{N}y_{p,i}\log(\hat{y}_{p,i}) + (1 - y_{p,i})\log(1 - \hat{y}_{p,i}).
    \label{eq:bce}
\end{equation}
As well, we compute the MSE of violence attributes,
\begin{equation}
    \mathcal{L}_{\text{Violence}}(\theta_S) = \frac{1}{N} \sum_{i=1}^{N}(\hat{v_i}-v_i)^2.
\end{equation}
And the binary cross-entropy over the visual attributes,
\begin{equation}
    \mathcal{L}_{\text{Attributes}}(\theta_S) = -\frac{1}{N}\sum_{i=1}^{N}\sum_{j=1}^{10}{(y^{(i)}_{a,j}\log(\hat{y}^{(i)}_{a,j}) +
    (1 - y^{(i)}_{a,j})\log(1 - \hat{y}^{(i)}_{a,j}))}.
\end{equation}
For the final hybrid loss, weighted by {$\alpha_1$, $\alpha_2$, $\alpha_3$},
\begin{equation}
    L_{\text{Downstream}}(\theta_S) = \alpha_1 \mathcal{L}_{\text{Protest}}+ \alpha_2\mathcal{L}_{\text{Violence}}  + \alpha_3 \mathcal{L}_{\text{Attributes}}. \label{eq:downstream}
\end{equation}
Following \citet{UCLA_protest}, we use a ResNet50 to classify each image for protest and visual attributes and to estimate the overall violence level, enabling assessment of synthetic data utility in downstream analysis. We optimize using mini-batch gradient descent, optimizing over the average loss per mini-batch.

\subsection{Attack Analysis}
\label{subsec:attack}
\paragraph{Threat Model} We consider a \emph{threat model} where only the downstream model, trained with synthetic samples, is released publicly. This threat model reflects common deployment scenarios in which only trained analytic models are released, while data generation pipelines and training data remain private. In this setting, the adversary does not have access to the generative model, the training data used to produce synthetic samples, or the synthetic samples themselves. As with previous work~\cite{jagielski2020auditing,lu2022general,Izzo2022ProvableMI,andrew2024oneshot}, our analysis uses membership inference as an empirical measure of privacy, serving as an auditing tool to assess potential leakage and to characterize relative privacy risk. To characterize privacy risk under different adversarial capabilities, we evaluate membership inference attacks in both black-box and white-box scenarios, reflecting best- and worst-case access levels.

\paragraph{Black-Box Attack} In the \emph{black-box attack}, the adversary has no knowledge, and can only query the model for access to information. We consider the attack formulated by \citet{salem2018mlleaks}. We query the downstream model, gathering the softmax outputs for all a large pool of inputs, including potential training data. Gathering the highest class posterior probability for each sample, we experiment to find an optimal threshold value for binary classification. This exploits the model overfitting directly, assuming that the model will be more confident for images contained in the training dataset. Given the threshold value, public image data $\mathbf{x}$, and the queried model $\mathcal{C}$, we classify samples as members of the training set using this formulation:

\begin{equation}
\text{black-box}(x, \text{threshold}) = \begin{cases} 
                \text{True}, & \text{if } \text{max}(\mathcal{C}(\mathbf{x})) \geq \text{thresh} \\
                \text{False}, & \text{otherwise}
                                        \end{cases}.
\end{equation}

\paragraph{White-Box Attack}We frame the white-box attack as a worst-case analysis of privacy risk. The adversary has full access to the released downstream model, including its architecture, hyperparameters, loss function, and training algorithm. This setting captures powerful insider or institutional adversaries and provides an upper bound on potential privacy leakage. Although they lack access to the generative model or synthetic samples, we assume they possess a small subset of the original private training data for supervised training. To assess the robustness of our method under this setting, we implement the supervised white-box attack proposed by \citet{Nasr_2019}. The attacker extracts features from the target model’s forward and backward passes, including hidden activations, output logits, loss values, gradients with respect to layer weights, and labels. These features are fed into an encoder network composed of CNN and fully connected layers, trained to predict whether each input sample was part of the target model’s training set. The encoder outputs a membership score optimized to approximate the probability of each data point being a member of the training set.

\subsection{Bias Analysis}
Generative modeling introduces the risk of amplifying or redistributing biases present in the training data. To assess this risk, we conduct a bias analysis of the proposed framework at both the generative and downstream task levels.

At the generative level, we analyze demographic distributions in the synthetic images using facial attribute analysis. Specifically, we compare the distributions of age, gender, and race attributes in the generated data against those observed in the original dataset, assessing whether certain demographic groups are under- or over-represented after synthesis.

At the downstream level, we evaluate bias in model predictions by measuring \emph{statistical parity} across demographic subgroups. We examine whether predicted protest presence, estimated violence levels, and visual attribute predictions differ systematically across sensitive attributes, indicating potential dependence between outcomes and demographic characteristics. These analyses are intended as empirical audits of representation and outcome disparities rather than causal fairness guarantees.

Because demographic imbalance is inherent to protest imagery, we conduct all bias comparisons relative to the original dataset and evaluation protocol introduced by \citet{UCLA_protest}. This ensures that observed disparities can be interpreted in the context of existing data biases rather than being attributed solely to the synthetic data generation or downstream modeling process.

\section{Experimental Evaluation}

\label{sec: Experiments}
\subsection{Datasets}
We primarily use the benchmark protest analysis dataset introduced by \citet{UCLA_protest} for training and testing. This dataset contains a total of 40,764 images, human-annotated for protest, violence level, and various visual attributes (sign, fire, police, etc.). The dataset contains over 10,000 protest positive images labeled for violence and attributes and over 20,000 protest negative samples. We retain the original train–test split defined by~\citet{UCLA_protest}, which withholds 8,153 images for testing. The training portion of this split is used for fine-tuning and downstream model training, as described in Section \ref{subsec:img_gen} and Section \ref{subsec:downstream} respectively. Table \ref{tab:datasets} shows the data distributions of all datasets used in our experiments. For each dataset, we resized samples to $256\times256\times3$ pixels, with bilinear sampling and normalized pixel inputs to 0-1.

\begin{table}[b]
    \centering
    \caption{Distributions of the datasets used in this work}
    \label{tab:datasets}
    \setlength{\tabcolsep}{1.5pt}
    \begin{tabular}{lccc@{\hspace{0.2cm}}c@{\hspace{0.2cm}}l}
    \toprule
       Dataset & Total  & Train & Test & Protest Samples & Attributes \\
    \midrule
       \citet{UCLA_protest} & 40,764 & 32,611 & 8,153 & 11,659   & Yes \\
       Ours & 40,764 & 32,611 & 8,153 & 11,659   & Yes \\
       VGKG & 106,817 & 106,817 & -- & 106,817 & No \\
       UCF-QNRF \cite{idrees2018composition} & 1,535 & 1,535 & -- &  -- & No \\
       JHU-CROWD++ \cite{sindagi2020jhucrowd} & 4,250 & 4,250 & -- & -- & No\\
       NWPU-Crowd \cite{Wang_2021} & 5,109 & 5,109 & -- & --  & No\\
    \bottomrule
       
    \end{tabular}
     
\end{table}

We use supplementary datasets for performing pre-training and organizing attacks, without mixing these data with the downstream training or test splits. We first organized a collection of crowd-counting datasets: UCF-QNRF \cite{idrees2018composition}, JHU-CROWD++ \cite{sindagi2020jhucrowd}, and NWPU-Crowd \cite{Wang_2021}. These combined gave a total of 11,015 images of crowds in various contexts, perspectives, and aspect ratios. Additionally, we leverage over 100,000 images from the VGKG protest dataset for pretraining. This dataset is an image subset from the GDELT Visual Global Knowledge Graph project, aimed at indexing news media \cite{GDELT}. Each sample contains an URL of an image confidently labeled as a protest event by the Google Cloudvision API~\cite{Google}. 

\subsection{Implementation Details}
\paragraph{Baseline Models} We compare the generative results to baseline generative models in the privacy and image generation domains: DP-GAN \cite{xie2018differentially} and WGAN-GP \cite{gulrajani2017improved}. For both, we used the classical DCGAN architecture, increasing the layer count to support $256\times256\times3$ pixels. Each was trained using the AdamW optimizer~\cite{Loshchilov2017DecoupledWD} with a learning rate of $1e-5$ and weight decay of $1e-4$ for the generator and discriminator. We trained each model for 100,000 iterations with a batch size of 1, allowing for 5 discriminator steps per iteration. For WGAN-GP \cite{gulrajani2017improved}, we used a gradient penalty coefficient of $\lambda = 10$. For DP-GAN \cite{xie2018differentially}, we used a weight clip of 0.01, and we applied various privacy constraints $\varepsilon ={1,10,50}$ to obtain a comprehensive evaluation of different utility and privacy trade-offs.

\paragraph{Image Generation} For StyleGAN3 \cite{karras2021aliasfree}, we trained for 50,000 kimg, with a batch size of 32 and $\gamma$ set to 2 for the R1 regularization in Eq. (\ref{eq:gen}). We also used the adaptive discriminator augmentation (ADA) procedure \cite{karras2021aliasfree}, adaptively applying pixel blitting, geometric, and color transformations to images before feeding them into the discriminator. For StyleGAN3 pre-training, we trained an unconditional model for each pre-training sets under identical conditions. During fine-tuning, we first randomly initialized weights associated with conditional inputs, then retrained the models under the same procedure. We generate a synthetic dataset with a one-to-one correspondence to the original dataset from \citet{UCLA_protest}, producing one synthetic image per each original label. This allows for direct comparison in downstream tasks while preserving dataset structure.

\begin{figure*}[t]
    \centering
    \includegraphics[width=\linewidth, trim=10 10 10 10, clip]{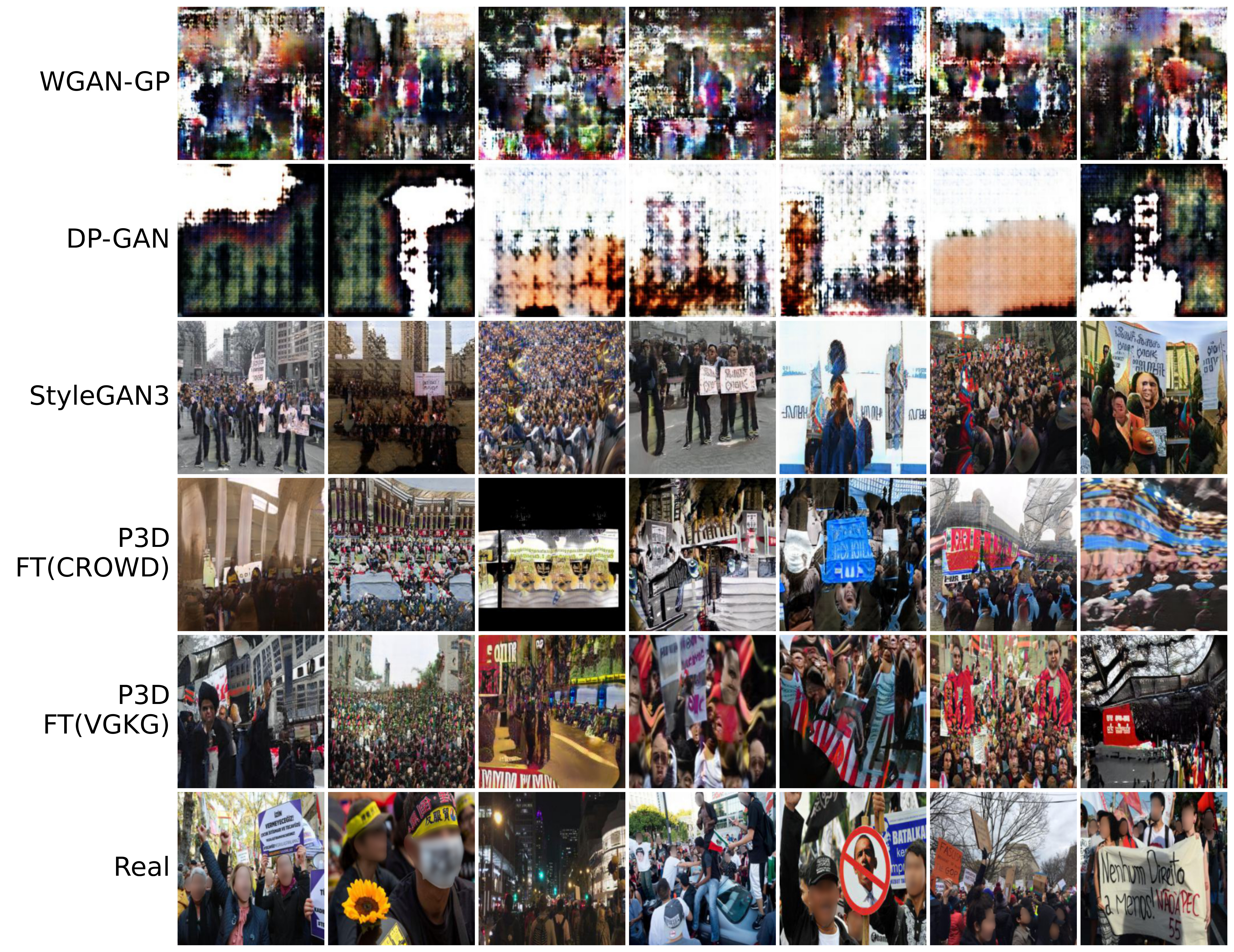}
     \caption{Qualitative examples of generated imagery for each generative model. We show the results of each generative model in each row and compare them to the real data sample (final row). Each column corresponds to a specific training set data entry, and it is conditionally generated using the corresponding annotation. Faces in the real images are blurred for privacy protection.  
     }
    \label{fig:qual}
    \Description{Samples from each generative model for protest imagery in this study.}
\end{figure*}

\paragraph{Downstream Protest Analysis} Following the approach of \citet{UCLA_protest}, we trained a ResNet-50 to predict protest classification labels, violence level normalized to 0-1, and visual attributes represented as binary labels of their presence in the scene. These visual attributes include: Sign, Photo, Fire, Police, Children, Group of 20 People, Group of 100 People, Flag, Night, and Shouting. We performed transformations to the inputs as in \citet{UCLA_protest}, which include a random resized crop of $224\times224$ pixels, a random rotation between $[-30,30]$ degrees, color jittering, and lighting noise. We trained our downstream model for 100 epochs on synthetic imagery, using SGD, a batch size of 32, and a learning rate of $0.002$. For the downstream loss (\ref{eq:downstream}), we set the hyper-parameters ${\alpha_1, \alpha_2, \alpha_3}$ $= {1,10,5}$ respectively. To compare baseline private model performance,  we trained DP versions of the same model using DP-SGD~\cite{abadi2016dpsgd} at $\varepsilon ={1, 10}$, with the only change being the use of GroupNorm~\cite{Wu2018GroupN} instead of BatchNorm layers.

\paragraph{Privacy Evaluation} To perform attacks, we compiled a dataset of 10,000 protest-positive images, with 5,000 images sourced from the \citet{UCLA_protest} dataset and 5,000 from VGKG. We performed an 80/20 split for training and testing. Our downstream model for protest analysis was used as the victim, and the adversary was modeled to have query access only. For the black-box attack, we implemented the simple threshold routine in Section \ref{subsec:attack}. For the white-box attack, we implemented a supervised membership inference method from \citet{Nasr_2019}, where the adversary is granted full access to the victim model's internals. To reduce computational cost and focus on salient information, we extracted features from only the final 10 layers of the target model. These features included per-layer hidden activations, per-layer gradients, loss, and true label with respect to a single input.

The white-box attack model's encoder is composed of both convolutional neural network (CNN) and multilayer perceptron (MLP) components. Different input feature types are routed through either CNNs (for structured multi-layer tensors like gradients) or MLPs (for scalar or low-dimensional features such as hidden layers or loss). The outputs of all components are concatenated into a single embedding vector, which is then passed to a final MLP for binary membership prediction. We trained the attack model using the Adam optimizer with a learning rate of $1\mathrm{e}{-4}$, a batch size of 32, and for 25 epochs.

\paragraph{Bias Evaluation}
As labels for individuals are not provided in the \citet{UCLA_protest} dataset, we employed a two-stage pipeline to infer sensitive attribute information. First, we used RetinaFace \cite{RetinaFace} to perform face localization across the protest datasets. Subsequently, we applied a state-of-the-art facial attribute detection method using a pre-trained VGG-Face model \cite{lightface2021}, trained on the FairFace dataset \cite{fairface_2021}, to extract demographic attributes. Classification scores were collected on both training and test data, using a confidence threshold of $\geq 0.5$ to ensure reliable face detections. We developed two experiments to evaluate bias in both the generative and downstream tasks: 
\begin{enumerate}
    \item Gathering facial attributes across synthetic and real training data to compare demographic representation. We sampled 2500 data samples and compared demographic distributions.
    \item Gathering facial attributes across the \citet{UCLA_protest} test dataset, we compared downstream protest analysis detection rates of our method within demographic subgroups.
\end{enumerate}
Since the model was trained on the FairFace \cite{fairface_2021} dataset, we expect some degradation in performance due to the domain shift to the protest imagery. Therefore, to validate robust performance, we have included an ablation study on 100 manually labeled protest images, as detailed in Appendix~\ref{sec:facial_analysis_ablation}.

\paragraph{Machine Configuration} This work was implemented in PyTorch, and models were trained using two NVIDIA A4000 GPUs, with 16 GB of VRAM each. We used a machine with an Intel Xeon(R) w7-2475X, 128GB of RAM, and 5TB of disk space to process and load data.

\subsection{Evaluation Metrics}
We evaluate the proposed framework across four dimensions: image quality, downstream analytical utility, privacy risk, and bias. To measure image quality, we used Frechet Inception Distance (FID)~\cite{heusel2018gans} and Kernel Inception Distance (KID)~\cite{bińkowski2021demystifying}, which measure the distance between feature distributions collected from the pre-trained Inception V3. Using the means $\{\mu_{\text{real}},\mu_{\text{synth}}\}$ and covariance matrices $\{C_{\text{real}},C_{\text{synth}}\}$ of the computed features $\{f_\text{real},f_\text{synth}\}$ in the deepest layer of Inception V3, FID measures the 2-Wasserstein distance between the real and synthetic feature distributions as follows: $\text{FID} = \| \mu_{\text{real}} - \mu_{\text{synth}} \|^{2} + \text{Tr}(C_{\text{real}} + C_{\text{synth}} - 2 \sqrt{C_{\text{real}}C_{\text{synth}}})$ where $Tr$ denotes the trace of a matrix. Like FID, KID uses Inception V3 to generate high level feature distributions of the data to compute the maximum mean discrepancy (MMD) as follows:  $ \text{KID} = \text{MMD}^2(f_{\text{real}}, f_{\text{synth}}) = \|\mu_{\mathcal{F}}(f_{\text{real}}) - \mu_{\mathcal{F}}(f_{\text{synth}})\|^2_{\mathcal{F}}$ where $ \mathcal{F}$ represents the reproducing kernel Hilbert space (RKHS) induced by a polynomial kernel. 

\begin{table}[b]
    \centering
    \caption{Quantitative evaluation of generative model performance. Visual quality and diversity metrics include FID, IS, and KID. For DP-GAN \cite{xie2018differentially} we show the results at varying privacy levels ${\epsilon = 1, 10, 50}$ as described in Eq.~(\ref{eq:diff_p}).}
    \label{tab:gen-quan}
    \setlength{\tabcolsep}{10pt}
    \begin{tabular}{l c c c} 
        Model & FID $\downarrow$ & IS $\uparrow$ & KID $\downarrow$ \\
        \toprule
        WGAN-GP \cite{gulrajani2017improved} & 303.524 & 1.736 & 0.343 \\  
        DP-GAN \cite{xie2018differentially} ($\epsilon = 1$) &  290.444 &  2.416  & 0.258 \\
        DP-GAN \cite{xie2018differentially} ($\epsilon = 10$) &  325.381 &  1.699  & 0.341 \\
        DP-GAN \cite{xie2018differentially} ($\epsilon = 50$) &  330.144 &  1.909  & 0.345 \\ 
        Ours (StyleGAN3) & 15.832 & 6.508 & 0.007 \\
        Ours FT (Crowd) & 40.263 & 5.827 & 0.024   \\ 
        Ours FT (VGKG) & 19.765 & 6.282 & 0.009 \\ 
        \bottomrule
    \end{tabular}
\end{table}

In addition, we used the Inception Score (IS)~\cite{salimans2016improved} to measure entropy and diversity of extracted features. Given synthetic image $\mathbf{x}$, we extracted the conditional probability distribution $P(\mathbf{y}|\mathbf{x})$ and marginal class distribution $P(\mathbf{y})$ using the features of Inception V3. The IS can then be calculated as follows: $\text{IS} = \exp\left(\mathbb{E}_x \left[ D_{\text{KL}}\left( P(\mathbf{y} | \mathbf{x}) \| P(\mathbf{y}) \right) \right]\right)$ where $D_{\text{KL}}$ is the Kullback-Leibler divergence. We report IS as a supplementary measure of diversity and classifier confidence, while relying primarily on FID and KID for assessing similarity to real data. We compute FID, KID, and IS using 50,000 generated images, as these metrics are known to suffer from poor accuracy given small datasets. For the downstream task, we assessed the area under the receiver operator characteristic curve (AUC-ROC) for each class label and the correlation of predicted and true violence levels. For the attack models, we measured the efficacy of identifying train samples using accuracy, precision, recall, and AUC-ROC. Lastly, to evaluate bias in downstream task, we calculate the pairwise statistical parity differences (SPD) between groups within each protected attribute. For groups \(i\) and \(j\), SPD is defined as the difference in positive outcome rates: $ \text{SPD}_{i,j} = P(\hat{Y} = 1 \mid A = i) - P(\hat{Y} = 1 \mid A = j)$, where \(\hat{Y}\) is the predicted label and \(A\) denotes group membership within a protected attribute (e.g., age, race, gender). A value of \(\text{SPD} = 0\) indicates parity between groups. These measures are used for descriptive auditing and do not imply normative or causal notions of fairness.

\subsection{Results and Discussion}

\paragraph{Quality} 
In protest imagery, visual coherence is critical, as misrendered crowds, police presence, or context can distort downstream interpretations of collective action and violence. In Table \ref{tab:gen-quan}, we quantitatively measure the quality and diversity of imagery from each generative model. For the dataset in \cite{UCLA_protest}, StyleGAN3 without pre-training achieves the best performance under FID, KID, and IS out of all the methods we have tested.Notably, our fine-tuning strategy has shown not to improve the overall quality of the synthetic imagery. This could be due to differences in dataset distributions between our pretraining sets and the target dataset, such as the lack of non-protest samples in pre-training. Our generations in Fig. \ref{fig:qual} show that WGAN-GP \cite{gulrajani2017improved} and DP-GAN \cite{xie2018differentially} are unable to reproduce the complex visual distributions present in protest imagery. We reason that the architectures used in these methods do not scale well to high resolution. Poor visual fidelity risks misrepresenting protest dynamics and introducing misleading cues into downstream analysis, making such models unsuitable for responsible deployment. Therefore, we exclude these models from the downstream task evaluation, rather than reporting potentially misleading results. Additional qualitative samples from the models can be found in Appendix~\ref{sec:additional_synthetic_samples}.

\begin{figure*}[t]
    \centering
    \subfloat[Protest recognition]{%
        \includegraphics[width=0.45\linewidth]{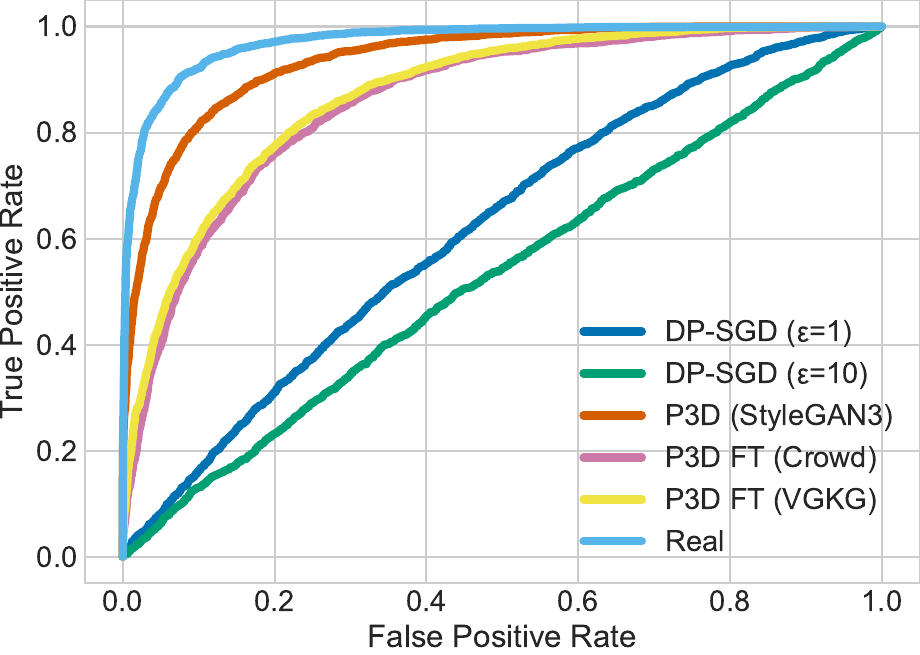}
        \label{fig:protest_recognition}
    }
    \hfill
    \subfloat[Violence prediction]{%
        \includegraphics[width=0.45\linewidth]{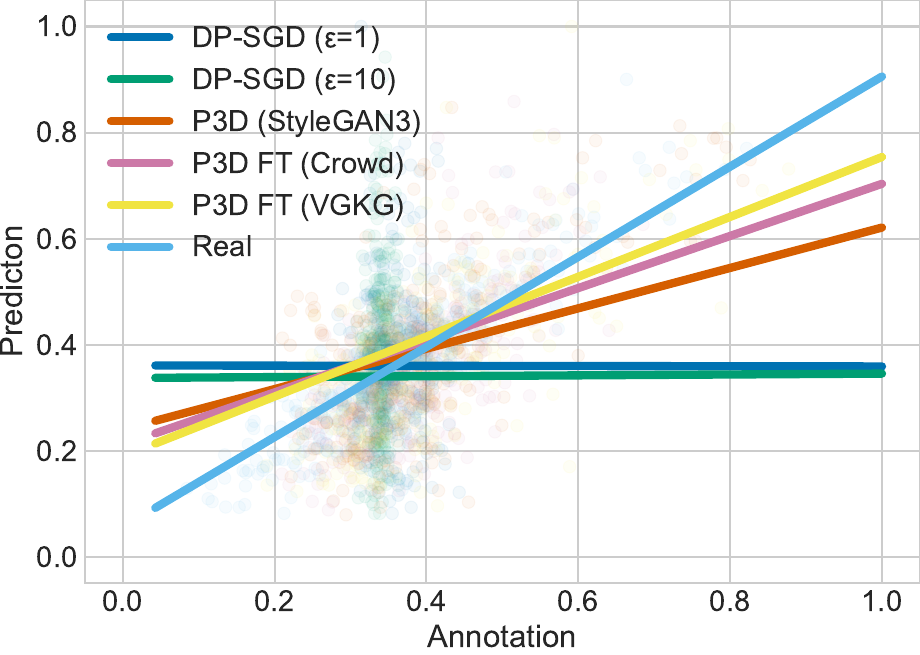}
        \label{fig:violence_prediction}
    }
    \caption{Downstream protest and violence prediction performance: (a) We report the receiver operating characteristic curve for downstream models trained on different synthetic data, as well as DP-SGD trained on real data. (b) Violence score performance is reported. We plot a scatter plot of 1000 samples from predicted violence on the test set and compute the line of best fit for each model using all test data samples.}
    \label{fig:both_plots}
    \Description{Protest and Violence prediction accuracy of downstream models.}
\end{figure*}

\begin{figure*}[t]
    \centering
    \includegraphics[width=0.95\linewidth]{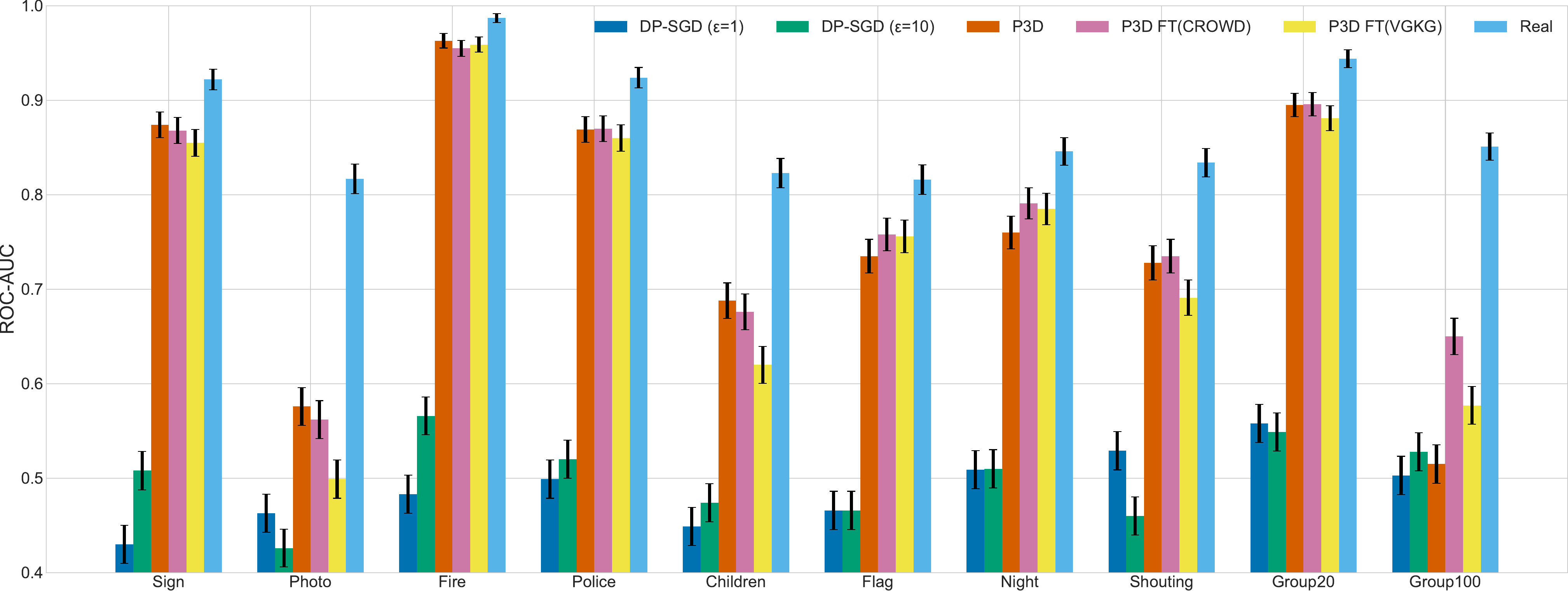}
    \caption{AUC-ROC per attribute of downstream classifiers on the UCLA test set when trained on the different conditionally generated data. Error bars represent 95\% confidence intervals computed using Wilson score.}
    \label{fig:att}
    \Description{Attribute prediction accuracy of downstream models.}
\end{figure*}

\paragraph{Downstream Task}
For the downstream task, we find our method substantially outperforms traditional DP-SGD applied to the downstream model at low privacy levels ($\varepsilon = 10$). These results suggest that high-utility downstream performance can be achieved without direct exposure to sensitive real images during downstream training. For protest accuracy, StyleGAN3 without pre-training serves as the most informative data to our downstream model. Fig. \ref{fig:both_plots}(a) shows the computed ROC curves for each model, showing increased performance over our pre-training strategy. For fine-tuned models, we observe mixed results in the downstream performance, likely due to the absence of conditioning during fine-tuning. In Fig. \ref{fig:both_plots}(b) models pre-trained using VGKG performed best when predicting violence scores. This may reflect increased exposure to violent visual patterns in the pre-training data. Lastly, attribute scores remained strongly represented in the generated imagery. We observe a small decrease from real training performance for most models. Fig \ref{fig:att} details the accuracy scores for all attributes. In particular, we notice a bigger gap in performance for categories 'Group20' and 'Group100' between models pre-trained with and without crowd counting data. These disparities motivate the bias analysis presented in the following section, where we examine whether such performance gaps correlate with demographic representation.

\paragraph{Privacy Analysis}
Results of the black-box and white-box attack can be found in Table \ref{tab:black_box} and Table \ref{tab:white_box} respectively. We compare against models trained on the real data, with and without using DP-SGD. As DP-SGD models exhibit low confidence in their softmax outputs, these outputs are much less exploitable. This behavior reflects reduced model confidence rather than task-based resistance to inference: the DP-SGD–trained target model produces near-uniform predictions, leaving the attack model with no discriminative signal and resulting in trivial membership decisions. Notably, even with a neutral threshold of 0.5, black-box attacks on models trained with DP-SGD essentially return all testing samples as members, resulting in a recall of 1.0 and precision around 0.5. Similarly, in the white-box attack, only positives were predicted, resulting in a precision and recall of 0.5 and 1 respectively. In the black-box setting, our method consistently exhibits lower attack accuracy, precision, and recall compared to models trained on real data. 

\begin{table}[h]
    \centering
    \caption{Reported accuracy, precision, and recall scores of classification using the block-box attack model. We experimented with softmax thresholds of $[0.95,0.99]$, and $[0.50]$ for DP-SGD due to its low confidence.}
    \setlength{\tabcolsep}{4pt}
    \begin{tabular}{lc cccc}
    \toprule
       Method  & \phantom{a} & Threshold & Accuracy & Precision & Recall \\
    \midrule
       \multirow{2}{*}{Real}
        && 0.950 & 0.592 & 0.563 & 0.871 \\
        && 0.990 & 0.621 & 0.627 & 0.616 \\
        \midrule
       \multirow{1}{*}{DP-SGD($\varepsilon=10$)}  
        && 0.500 & 0.506 & 0.506 & 1.000 \\
        \midrule
       \multirow{1}{*}{DP-SGD($\varepsilon=1$)}  
        && 0.500 & 0.506 & 0.506 & 1.000 \\
    
    \midrule
       \multirow{2}{*}{Ours}  
        && 0.950 & 0.532 & 0.540 & 0.505 \\
        && 0.990 & 0.539 & 0.617 & 0.233 \\

    \bottomrule\\
       
    \end{tabular}
    
    \label{tab:black_box}
\end{table}

In the white-box attack, we find attacks on our framework exhibit much lower recall, and much higher log-loss. This suggests that synthetic training data weakens the memorization patterns typically exploited by white-box membership inference attacks. These results illustrate an empirical privacy–utility trade-off: while DP-SGD provides stronger empirical resistance to membership inference, it does so at the cost of image fidelity and downstream performance, whereas our framework achieves substantially higher utility with weaker, but non-trivial, privacy protection. We emphasize that membership inference serves as an empirical proxy for privacy risk and does not constitute a formal guarantee.

\begin{table}[h]
    \centering
    \caption{Reported accuracy, precision, recall, and AUC-ROC scores of classification using the white-box attack model. Our results show a considerable advantage in the worse case attack scenario, compared to training on real data.}
    \setlength{\tabcolsep}{4pt}
    \begin{tabular}{lc cccccc}
    \toprule
       Attack Model & \phantom{a} & Method & Accuracy & Precision & Recall & AUC-ROC & Log-Loss \\
    \midrule
       \multirow{4}{*}{\citet{Nasr_2019}} && Real & 0.703 & 0.674 & 0.786 & 0.794 & 1.908 \\
                    && DP-SGD($\varepsilon=10$)  & 0.500 & 0.500 & 1.000 & 0.500 & 0.693 \\
                    && DP-SGD($\varepsilon=1$)  & 0.500 & 0.500 & 1.000 & 0.500 & 0.693 \\
                    && Ours & 0.679 & 0.689 & 0.654 & 0.733 & 2.4579 \\
    \midrule
       
    \end{tabular}
    \label{tab:white_box}
\end{table}

\begin{figure*}[h]
  \centering
  
  \begin{subfigure}{\textwidth}
    \centering
    \includegraphics[width=\textwidth]{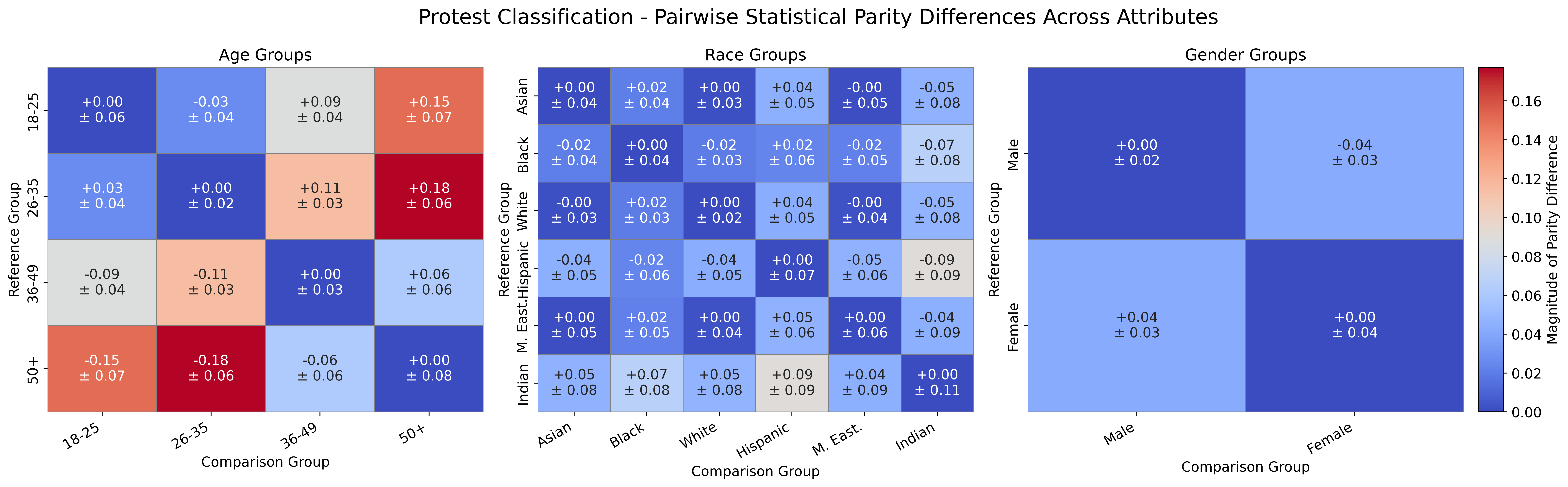}
    \label{fig:subfig1}
  \end{subfigure}
  
  
  \begin{subfigure}{\textwidth}
    \centering
    \includegraphics[width=\textwidth]{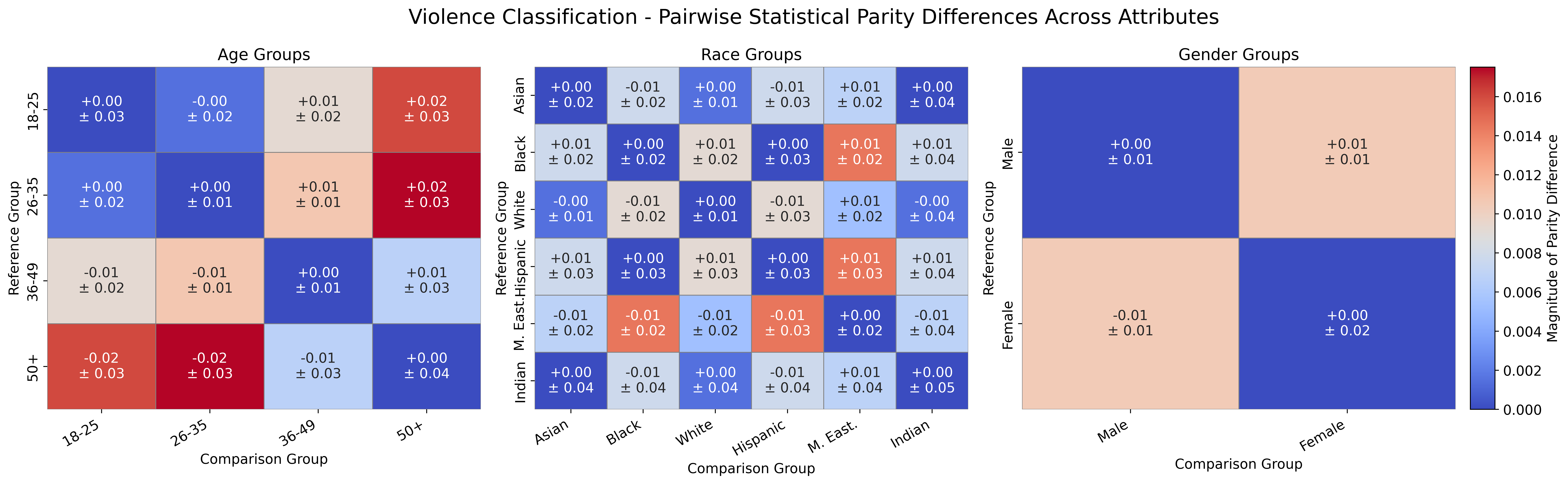}
    \label{fig:subfig2}
  \end{subfigure}
  
  \caption{ Pairwise statistical parity differences for predicted outcomes across sensitive attributes: age, race, and gender. Each heatmap cell compares the prediction rates between two demographic groups (Comparison Group vs Reference Group) for a given attribute. The values represent the difference in prediction rates (Comparison Group – Reference Group), with 95\% confidence intervals shown below each value. These intervals are computed using Wilson score. Darker red colors indicate larger disparities in prediction rates, highlighting potential fairness concerns.}
  \label{fig:bias}
  \Description{Pairwise statistical parity differences for predicted outcomes of our resulting model. Plotted as confusion matrices.}
\end{figure*}

\begin{figure*}[h]
\begin{center}
\subfloat[\label{fig:ages}]{
    \includegraphics[width=0.31\linewidth]{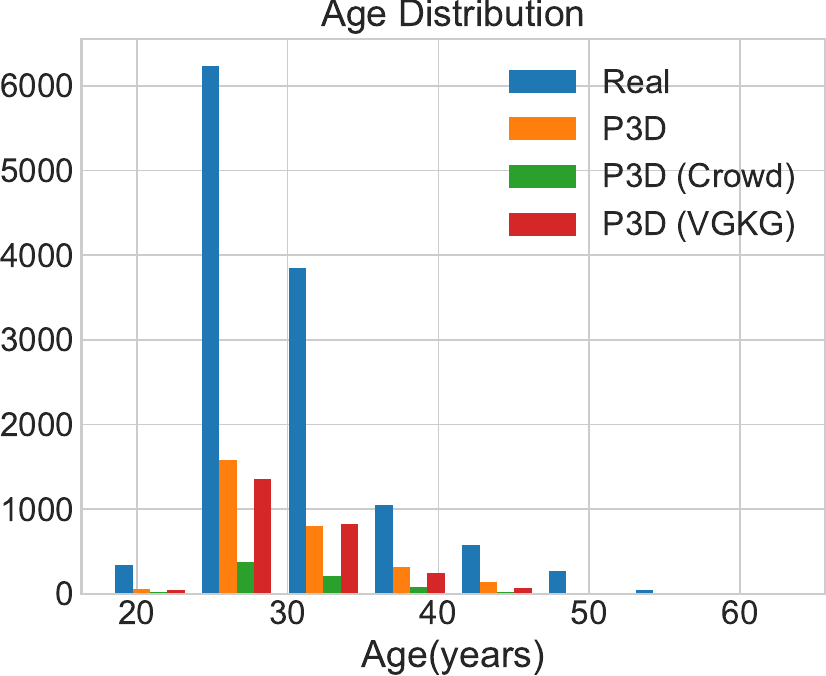}
}\hfill
\subfloat[\label{fig:genders}]{
    \includegraphics[width=0.31\linewidth]{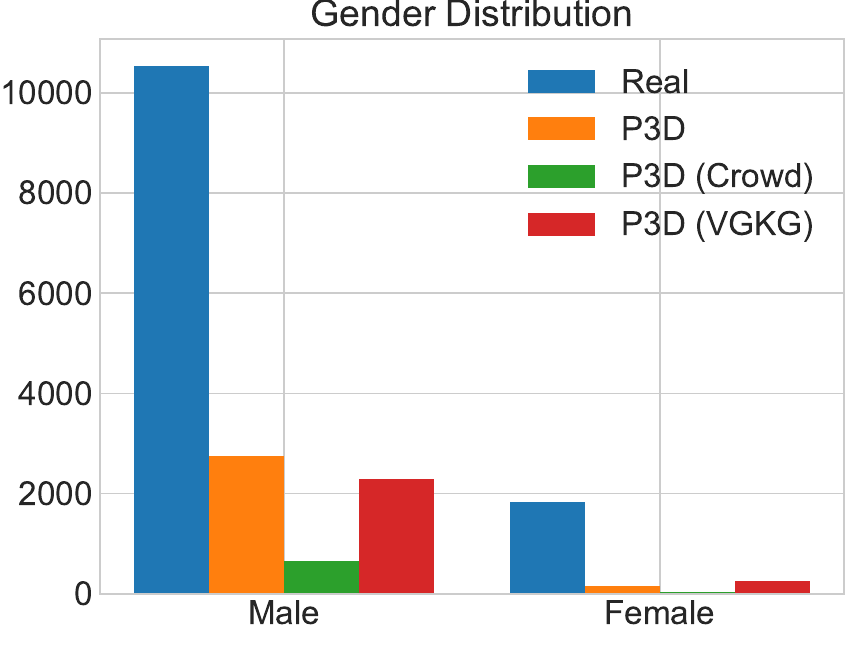}
}\hfill
\subfloat[\label{fig:race}]{
    \includegraphics[width=0.31\linewidth]{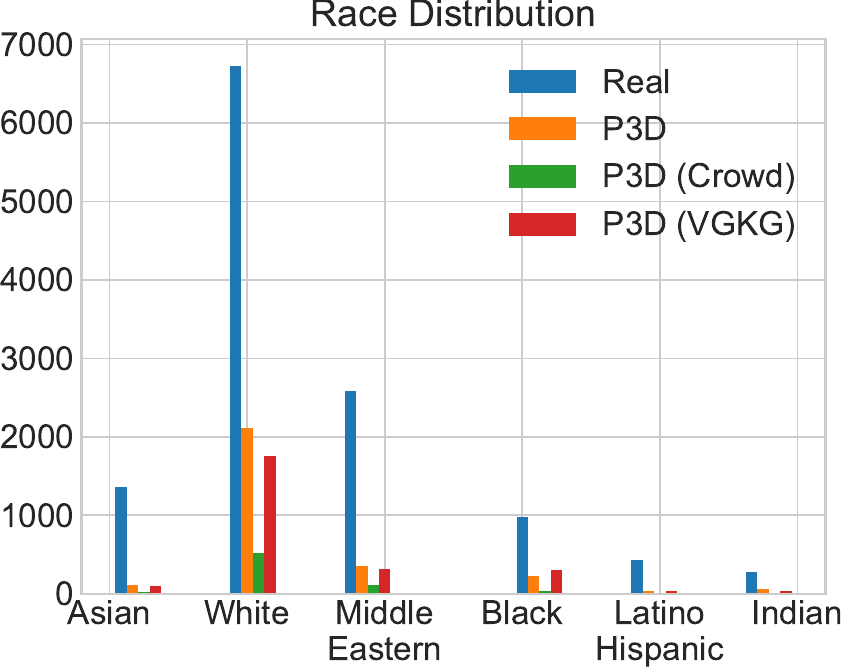}
}
\caption{Face Attribute distributions from synthetic and real datasets, assessing (a) Age, (b) Gender, and (c) Race. We randomly sample 2500 images from each training set, and extract the facial attributes of detected faces with $\geq0.5$ confidence using a pre-trained VGG-Face model \cite{lightface2021} trained on the FairFace dataset \cite{fairface_2021}.}
\Description{Face attribute statistics collected from both real and synthetic datasets.}
\label{fig:face_att}
\end{center}%
\end{figure*}

\paragraph{Bias Evaluation}
The results of our facial attribute detection for age, gender, and race in the training datasets are shown in Fig. \ref{fig:face_att}. Our bias analysis shows consistent detected attributes across generated faces between our synthetic and reference data. Interestingly, we notice when pre-training using crowd data face detection rates are substantially lower. In addition, when using VGKG, we observe an increase in the presence of underrepresented groups such as individuals aged 30–40, Black individuals, and women. For the downstream task, we report the statistical parity difference across subgroups in Fig. \ref{fig:bias}. Observing attributes with the highest risk, protest and violence, we observe higher predicted protest rates in the model outputs for female and Indian subgroups and a slightly higher violence detection rates for races of darker skin tone. Though, some of these sub-group differences can be attributed to low sample size. Age groups report the highest disparity in both protest and violence predictions, which may partially stem from inaccuracies in the face analysis model, particularly in estimating age under challenging conditions. While modest, these differences highlight the need for explicit fairness interventions to ensure equitable performance across age demographics. Additionally, we performed an ablation study to evaluate face-analysis accuracy; the most common failure modes involved misclassifications under poor lighting, partial occlusions and low resolution, all frequent in protest settings. Details of the ablation study are provided in Appendix \ref{sec:facial_analysis_ablation}.

\paragraph{Limitations \& Deployment Risks}
While our approach demonstrates promising results, certain limitations and potential risks must be acknowledged. The benefits of pretraining on external datasets were modest in our experiments, and the trade-off between increased training complexity and uncertain generalizability should be carefully considered. From a privacy standpoint, although synthetic images reduce direct exposure to real individuals, risks such as latent space inversion or re-identification remain possible, especially if auxiliary information is compromised. These vulnerabilities highlight the need for safeguards during deployment and access control, alongside awareness of dual-use risks, such as potential misuse for surveillance or repression.

\paragraph{Compliance with Data Privacy Regulations}
While it remains an open question whether synthetic data fully satisfies legal standards for personal identifiable data protection under regulations such as GDPR and CCPA, our method conceptually aligns with the goals of these frameworks by reducing reliance on real user data. The synthetic data we generate contains no directly identifiable personal information by design, helping to mitigate privacy risks and supporting the broader principles of data protection, minimization, and responsible data stewardship.

\paragraph{Future Work}
We chose StyleGAN3 for its high-fidelity image synthesis and its superior control over input attributes, but recent diffusion models present a promising alternative for improving output diversity and scalability. Future work could explore diffusion-based methods and expanded conditional generation to better represent protest attributes and protected demographics in a balanced manner. Generating multiple synthetic samples per label may also boost data augmentation and model robustness. While we used ResNet-50 for consistency, evaluating architectures like EfficientNet or Vision Transformers could offer further insights into generalization across datasets and deployment settings. Additionally, evaluating privacy risks beyond membership inference attacks, such as model inversion and attribute inference, is an important direction for future work.

\section{Conclusion}
By replacing sensitive datasets with synthetic reproductions, our proposed method enables utilizable and privacy-aware downstream protest analysis. Our experiments demonstrate that GAN-based synthetic data can acheive a balance of privacy, fairness, and utility, surpassing baseline methods in capturing complex protest imagery features under the evaluated settings. Where traditional privacy-aware methods substantially degrade utility, we provide a practical and efficient framework that supports equitable and responsible analysis of protest imagery, with the goal of contributing to positive social outcomes. Our ongoing work is focused on larger-scale validation and comprehensive risk assessment, including expanded fairness evaluation, to ensure robust and ethical deployment.



\bibliographystyle{ACM-Reference-Format}
\bibliography{bibliography}

\appendix

\section{Facial Analysis Ablation}
\label{sec:facial_analysis_ablation}

To assess the performance of our face analysis pipeline on protest imagery, we conducted a manual labeling experiment on a randomly sampled subset of 100 protest images from our dataset. Each image was annotated for the presence of visible human faces and, where applicable, labeled with estimated demographic attributes: age group, perceived race, and gender.

Two human annotators independently labeled each image, and disagreements were resolved through consensus to establish a ground truth. We then ran our face detection and analysis pipeline on the same images and compared the results against the manual annotations to evaluate system performance.

\begin{table}[h!]
\centering
\caption{Performance metrics for face detection and demographic attribute classification on 100 manually labeled protest images.}
\label{tab:face_analysis_metrics}
\begin{tabular}{lccccc}
\toprule
\textbf{Task} & \textbf{Accuracy} & \textbf{Precision} & \textbf{Recall} & \textbf{mIOU} & \textbf{mAP@0.5} \\
\midrule
Face Detection       & —        & —        & —        & 0.737    & 0.816   \\
Age Classification   & 36.2\%      & 0.405        & 0.291        & —      & —     \\
Race Classification  & 60.9\%      & 0.689        & 0.429        & —      & —     \\
Gender Classification&  72.5\%      & 0.381        & 0.308        & —      & —     \\
\bottomrule
\end{tabular}
\end{table}

This ablation confirms that while our pipeline achieves reasonable performance, caution should be taken when interpreting demographic attributes in complex, in-the-wild protest imagery. These results support our motivation for integrating fairness considerations and uncertainty estimation into downstream analyses.

\section{Additional Synthetic Protest Imagery}
\label{sec:additional_synthetic_samples}

\begin{figure*}[h]
    \centering
    \includegraphics[width=0.9\linewidth]{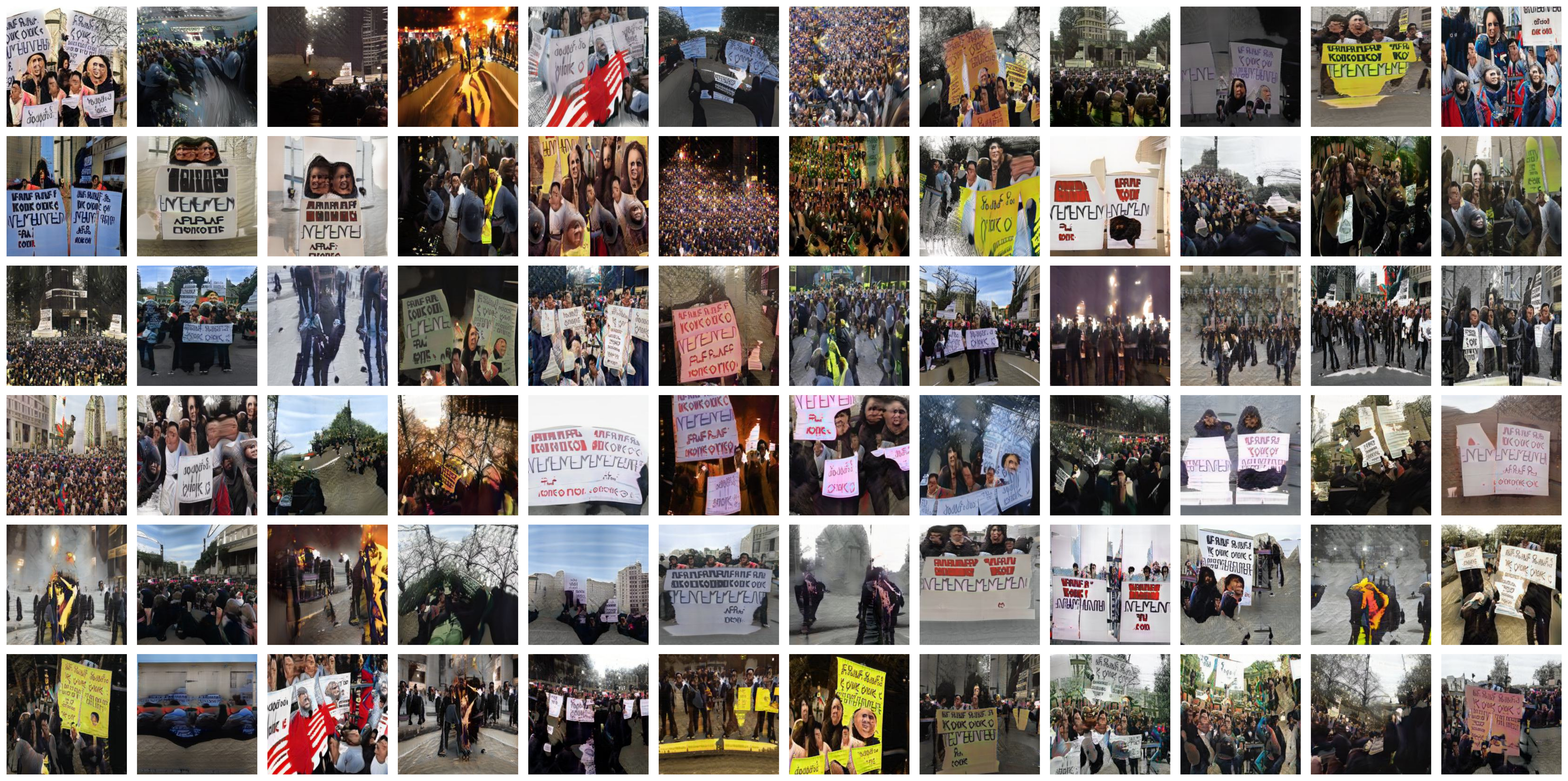}
    \caption{Additional samples of synthetic images}
    \label{fig:base_p3d_samples}
    \Description{Additional samples of synthetic images}
\end{figure*}

\begin{figure*}[h]
    \centering
    \includegraphics[width=\linewidth]{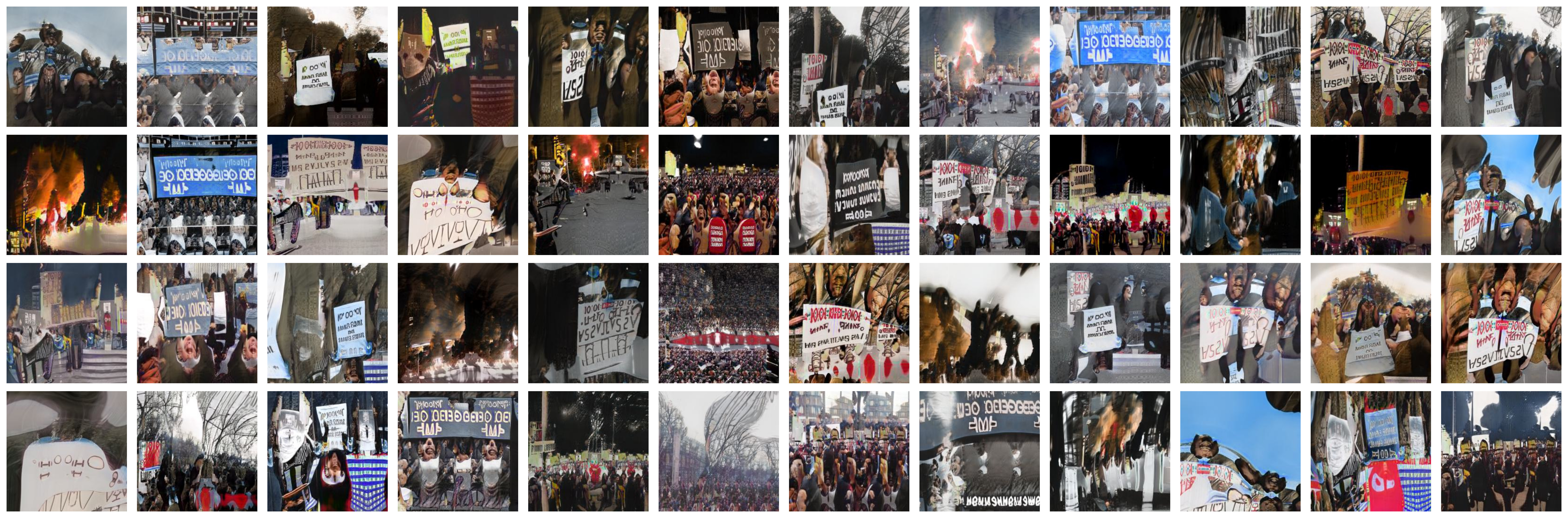}
    \caption{Additional samples of synthetic images pre-trained on crowd-counting datasets}
    \label{fig:cwdcnt_p3d_samples}
    \Description{Additional samples of synthetic images pre-trained on crowd-counting datasets}
\end{figure*}

\begin{figure*}[h]
    \centering
    \includegraphics[width=\linewidth]{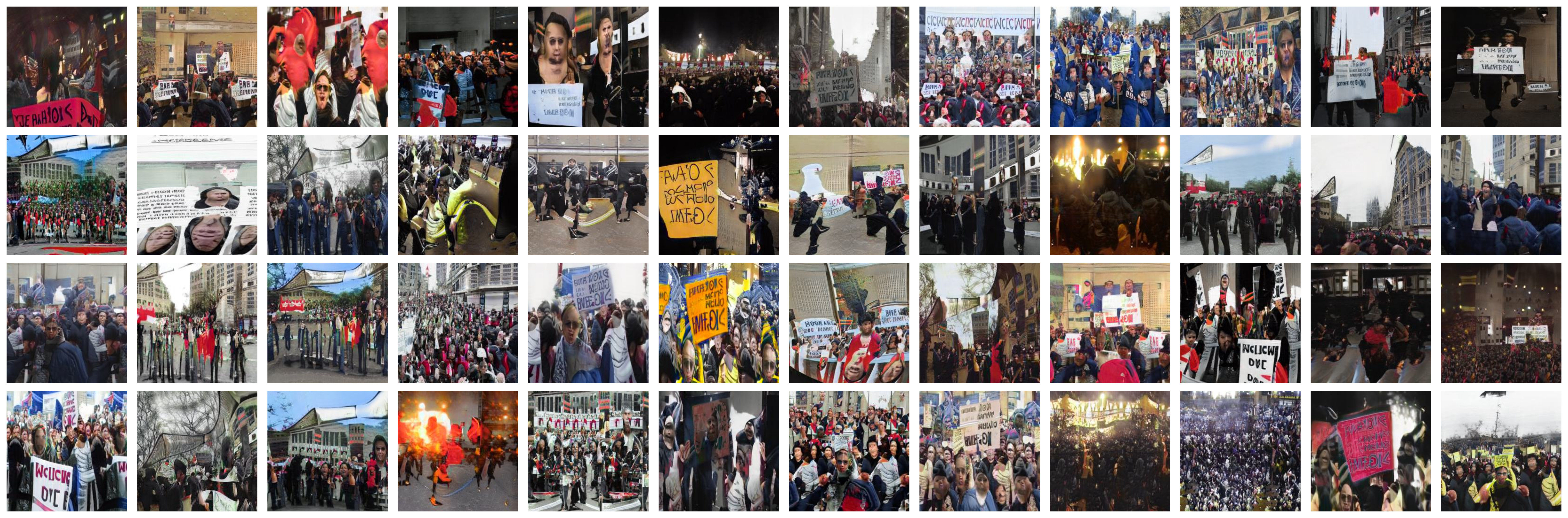}
    \caption{Additional samples of synthetic images pre-trained on the VGKG dataset}
    \label{fig:vgkg_p3d_samples}
    \Description{Additional samples of synthetic images pre-trained on the VGKG dataset}
\end{figure*}

This section presents additional synthetic protest images generated by our framework under various training settings:

\begin{itemize}
\item \emph{Base Model:} Figure~\ref{fig:base_p3d_samples} includes protest samples from the baseline model trained on \citet{UCLA_protest}.
\item \emph{Crowd-Counting Pretrained Model:} Figure~\ref{fig:cwdcnt_p3d_samples} shows protest samples from a model pretrained on crowd-counting data.
\item \emph{VGKG Pretrained Model:} Figure~\ref{fig:vgkg_p3d_samples} contains protest samples from a model pretrained on VGKG imagery.
\end{itemize}

These figures offer a qualitative overview of outputs under different pretraining regimes.

\end{document}